\documentclass[runningheads]{llncs}
\usepackage{graphicx}
\usepackage{amsmath,amssymb} 
\usepackage{color}
\usepackage[width=122mm,left=12mm,paperwidth=146mm,height=193mm,top=12mm,paperheight=217mm]{geometry}

\usepackage{epsfig}
\usepackage[acronym]{glossaries}
\usepackage{multirow}
\usepackage{caption}
\usepackage{float} 
\usepackage{subcaption}
\usepackage{adjustbox}
\usepackage{array}

\def\*#1{\mathbf{#1}}
\DeclareMathOperator*{\argmin}{arg\,min}

\newacronym{cnn}{CNN}{Convolutional Neural Network}
\newacronym{mse}{MSE}{Mean-Squared Error}
\newacronym{gan}{GAN}{Generative Adversarial Networks}
\newacronym{psnr}{PSNR}{Peak Signal-to-Noise Ratio}

\usepackage[pagebackref=true,breaklinks=true,letterpaper=true,colorlinks,bookmarks=false]{hyperref}
\usepackage[capitalise]{cleveref}

\graphicspath{{images}}

\newcommand*\samethanks[1][\value{footnote}]{\footnotemark[#1]}

\newcolumntype{R}[2]{%
    >{\adjustbox{angle=#1,lap=\width-(#2)}\bgroup}%
    l%
    <{\egroup}%
}
\newcolumntype{P}[1]{>{\centering\arraybackslash}p{#1}}

\begin{document}
\pagestyle{headings}
\mainmatter
\def\ECCV18SubNumber{***}  

\title{Frame Interpolation with Multi-Scale Deep Loss Functions and Generative Adversarial Networks}

\author{Joost van Amersfoort\thanks{These authors contributed equally to this work.} , Wenzhe Shi\samethanks , Alejandro Acosta, Francisco Massa, \\
Johannes Totz, Zehan Wang, Jose Caballero\samethanks \\
{\tt\small joost.van.amersfoort@cs.ox.ac.uk}\\
{\tt\small \{wshi, aacostadiaz, fmassa, jtotz, zehanw, jcaballero\}@twitter.com}
}
\institute{Twitter}

\titlerunning{FIGAN}

\authorrunning{J. van Amersfoort et al.}

\maketitle

\begin{abstract}
Frame interpolation attempts to synthesise frames given one or more consecutive video frames. In recent years, deep learning approaches, and notably convolutional neural networks, have succeeded at tackling low- and high-level computer vision problems including frame interpolation. These techniques often tackle two problems, namely algorithm efficiency and reconstruction quality. In this paper, we present a multi-scale generative adversarial network for frame interpolation (\mbox{FIGAN}). To maximise the efficiency of our network, we propose a novel multi-scale residual estimation module where the predicted flow and synthesised frame are constructed in a coarse-to-fine fashion. To improve the quality of synthesised intermediate video frames, our network is jointly supervised at different levels with a perceptual loss function that consists of an adversarial and two content losses. We evaluate the proposed approach using a collection of 60fps videos from YouTube-8m. Our results improve the state-of-the-art accuracy and provide subjective visual quality comparable to the best performing interpolation method at $\times 47$ faster runtime.
\end{abstract}

\section{Introduction}

Frame interpolation attempts to synthetically generate one or more intermediate video frames from existing ones, the simple case being the interpolation of one frame given two adjacent video frames. This is a challenging
problem requiring a solution that can model natural motion within a video, and generate frames that respect this modelling.
Artificially increasing the frame-rate of videos enables a range of applications. For example, data compression can be achieved by actively dropping video frames at the emitting end and recovering them via interpolation on the receiving end
\cite{sekiguchi2005lowcost}. Increasing video frame-rate also directly allows to
improve visual quality or to obtain an artificial slow-motion effect
\cite{twixtor,meyer2015phase,liu2017video}.

\begin{figure}
\centering
  \begin{subfigure}{0.295\textwidth}
  \centering
  		\includegraphics[width=\textwidth]{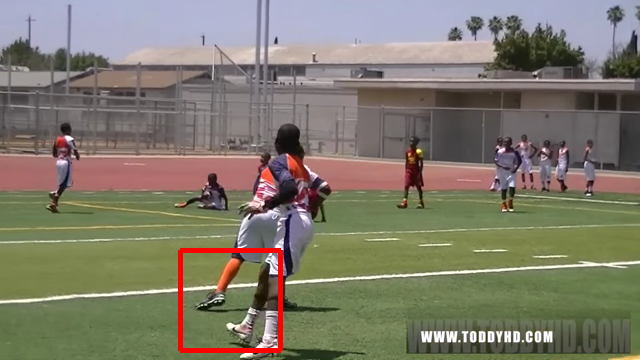}
       \caption{Original image}
  \end{subfigure}
  \hfill
  \begin{subfigure}{0.165\textwidth}
  		\includegraphics[width=\textwidth]{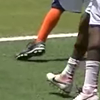}
       \caption{Original}
  \end{subfigure}
  \hfill
  \begin{subfigure}{0.165\textwidth}
  		\includegraphics[width=\textwidth]{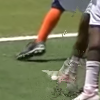}
       \caption{\cite{WulffCVPR2015}}
  \end{subfigure}
  \hfill
  \begin{subfigure}{0.165\textwidth}
  		\includegraphics[width=\textwidth]{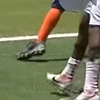}
       \caption{\cite{niklaus2017sepconv}}
  \end{subfigure}
  \hfill
  \begin{subfigure}{0.165\textwidth}
  		\includegraphics[width=\textwidth]{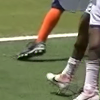}
       \caption{FIGAN}
  \end{subfigure}
  \caption{Visual example of frame interpolation results using PCA-layers \cite{WulffCVPR2015}, SepConv $\mathcal{L}_F$ \cite{niklaus2017sepconv}, and the proposed solution FIGAN, which combines a multi-scale neural network design with a perceptual training loss that surpasses state-of-the-art accuracy with real-time runtime. Visual quality is comparable to SepConv \cite{niklaus2017sepconv} while requiring $\times 3.24$ fewer computations.}
\end{figure}

Frame interpolation commonly relies on optical flow 
\cite{werlberger2011optical,raket2012motion,baker2011middlebury,liu2017video}. Optical flow relates consecutive frames in a sequence describing 
the displacement that each pixel undergoes from one frame to the next. One 
solution for frame interpolation is therefore to assume constant velocity in 
motion between existing frames and interpolate frames via warping. However, optical flow 
estimation is difficult and time-consuming, and a good illustration of this is that the average run-time per $480 \times 640$ frame of the top five performing methods of 2017 in the Middlebury benchmark dataset \cite{baker2011middlebury} is 1.47 minutes\footnote{Runtime reported by authors and not normalised by processor speed.}. Furthermore, there is in general no consensus on a single model that can 
accurately describe it. Different models have been proposed based on inter-frame 
colour or gradient constancy, but these are susceptible to failure in challenging 
conditions such as occlusion, illumination or nonlinear structural changes. As a 
result, methods that obtain frame interpolation as a derivative of flow 
suffer from inaccuracies in flow estimation.

Recently, deep learning approaches, and in particular \glspl{cnn}, have set up new state-of-the-art results across many computer vision problems, and have also 
provided new optical flow estimation methods. In \cite{fischer2015flownet,ilg2016flownet}, optical flow features are trained in a supervised setup mapping 
two frames to their ground truth optical flow field. Spatial transformer networks \cite{jaderberg2015spatial} allow an image to be spatially transformed as part of a differentiable network, learning a transformation implicitly in an unsupervised fashion, hence enabling frame interpolation with an end-to-end differentiable network \cite{liu2017video}. Choices in network design and training strategy can directly affect 
interpolation quality as well as efficiency. Multi-scale residual estimations 
have been repeatedly proposed in the literature \cite{ranjan2016optical,weinzaepfel2013deepflow,raket2012motion}, but only simple models based on colour constancy have 
been explored. More recently, training strategies have been proposed for 
low-level vision tasks to go beyond pixel-wise error metrics making use of more 
abstract data representations and adversarial training, producing visually more pleasing results \cite{johnson2016perceptual,ledig2016photo}. An example of this notion applied to frame interpolation networks is explored very recently in \cite{niklaus2017sepconv}.

In this paper we propose a real-time frame interpolation method that can 
generate realistic intermediate frames with high \gls{psnr}. It is the first 
model that combines the pyramidal structure of classical optical flow modeling 
with recent advances in spatial transformer networks for frame interpolation. 
Compared to naive CNN processing, this leads to a $\times 9,3$ 
speedup with a $2.38$dB increase in \gls{psnr}. Furthermore, to work around natural 
limitations of intensity variations and nonlinear deformations, we investigate 
deep loss functions and adversarial training. These contributions result in an interpolation model that is more expressive and informative relative to models based solely on pixel intensity losses as illustrated in \cref{table:soa} and \cref{fig:soa}.

\section{Related Work}

\subsection{Frame interpolation with optical flow}

The main challenge in frame interpolation lies in respecting object motion 
and occlusions such as to recreate a plausible frame that preserves structure 
and consistency of data. Although there has been work in frame 
interpolation without explicit motion estimation \cite{meyer2015phase}, the vast majority of methods relies on flow estimation as a description of motion across frames \cite{baker2011middlebury,raket2012motion,liu2017video}.

Let us define two consecutive frames with $I_0$ and $I_1$, their optical flow 
relationship can be formulated as
\begin{equation}
I_0(x, y) = I_1(x + u, y + v),
\end{equation}
where $u$, and $v$ are pixel-wise displacement fields for dimensions $x$ and 
$y$. For convenience, we will use the shorter notation $I(\Delta)$ to refer to 
an image $I$ with coordinate displacement $\Delta=(u, v)$, and write $I_0 = 
I_1(\Delta)$. Multiple strategies can be adopted for the estimation of 
$\Delta$, ranging from a classic minimisation of an energy functional given flow 
smoothness constraints \cite{horn1980flow}, to recent proposals employing neural 
networks \cite{fischer2015flownet}. Flow amplitude can vary 
greatly (eg. slow moving details versus camera panning), and in order to efficiently account for 
this variability flow can be approximated at multiple scales. Finer flow scales 
take advantage of estimations at lower coarse scales to progressively 
estimate the final flow in a coarse-to-fine fashion \cite{brox2004high,brox2011large,hu2016efficient}. Given an optical flow between two frames, an 
interpolated intermediate frame $\hat I_{0.5}$ can be estimated by projecting 
the flow at time $t=0.5$ and pulling intensity values bidirectionally from 
frames $I_0$ and $I_1$. A description of this interpolation mechanism can be 
found in \cite{baker2011middlebury}. 

\subsection{Neural networks for frame interpolation}

Neural network solutions have been proposed for the supervised learning of 
optical flow fields from labelled data \cite{fischer2015flownet,ilg2016flownet,niklaus2017adaconv,niklaus2017sepconv}. Although these have been successful and 
could be used for frame interpolation in the paradigm of flow estimation and 
independent interpolation, there exists an inherent limitation in that flow 
labelled data is scarce and expensive to produce. It is possible to work around 
this limitation by training on rendered videos where ground truth flow is known 
\cite{fischer2015flownet,ilg2016flownet}, although this solution is susceptible to 
overfitting synthetically generated data. An approach to directly interpolate 
images has been recently suggested in \cite{niklaus2017adaconv,niklaus2017sepconv} where large convolution kernels are estimated for each 
output pixel value. Although results are visually appealing, the complexity of these approaches has not been constrained to meet real-time runtime requirements.

Spatial transformers \cite{jaderberg2015spatial} have recently been used for unsupervised learning of optical flow by learning how 
to warp a video frame onto its consecutive frame \cite{ren2016unsupervised,yu2016back,ahmadi2016unsupervised}. In \cite{liu2017video} it is used to 
directly synthesise an interpolated frame using a \gls{cnn} to estimate flow 
features and spatial weights to handle occlusions. Although flow is estimated at 
different scales, fine flow estimations do not reuse coarse flow estimation like in the traditional pyramidal flow estimation paradigm, potentially indicating design inefficiencies.

\subsection{Deep loss functions}

Low-level vision problem optimisation often 
minimise a pixel-wise metric such as squared or absolute errors, 
as these are objective definitions of the distance between true data and its 
estimation. However, it has been recently shown how departing from pixel space 
to evaluate modelling error in a different, more abstract dimensional space can 
be beneficial. In \cite{johnson2016perceptual} it is shown how high dimensional features from 
the VGG network are helpful in constructing a training loss function that 
correlates better with human perception than \gls{mse} for the task of image 
super-resolution. In \cite{ledig2016photo} this is further enhanced with the use 
of \glspl{gan}. Neural network solutions for frame interpolation have been 
limited to the choice of classical objective metrics such as colour constancy \cite{liu2017video,niklaus2017adaconv}, but recently 
\cite{niklaus2017sepconv} has shown how perceptual losses can also be beneficial 
for this problem. Training for frame interpolation involving adversarial losses 
have nevertheless not yet been explored.

\subsection{Contribution}

We propose a neural network solution for frame interpolation that benefits from 
a multi-scale refinement of the flow learnt implicitly. The structural change to 
progressively apply and estimate flow fields has runtime implications as it 
presents an efficient use of computational network resources compared to the baseline as illustrated in \cref{table:design}. Additionally, we introduce a synthesis refinement module 
for the interpolation result inspired by \cite{ganin2016deepwarp}, which shows helpful in correcting reconstruction results. Finally, we propose a higher level, more expressive interpolation error 
modelling taking account of classical colour constancy, a perceptual loss and an 
adversarial loss functions. Our main contributions are:

\begin{itemize}
  \item A real-time neural network for frame interpolation.
  \item A multi-scale network architecture inspired by multi-scale optical flow 
estimation that progressively applies and refines flow fields.
  \item A reconstruction network module that refines frame synthesis results.
  \item A training loss function that combines colour constancy, a perceptual 
and adversarial losses.
  \end{itemize}

\section{Proposed Approach}
\label{proposed}

\begin{figure}[t]
  \includegraphics[width=\textwidth]{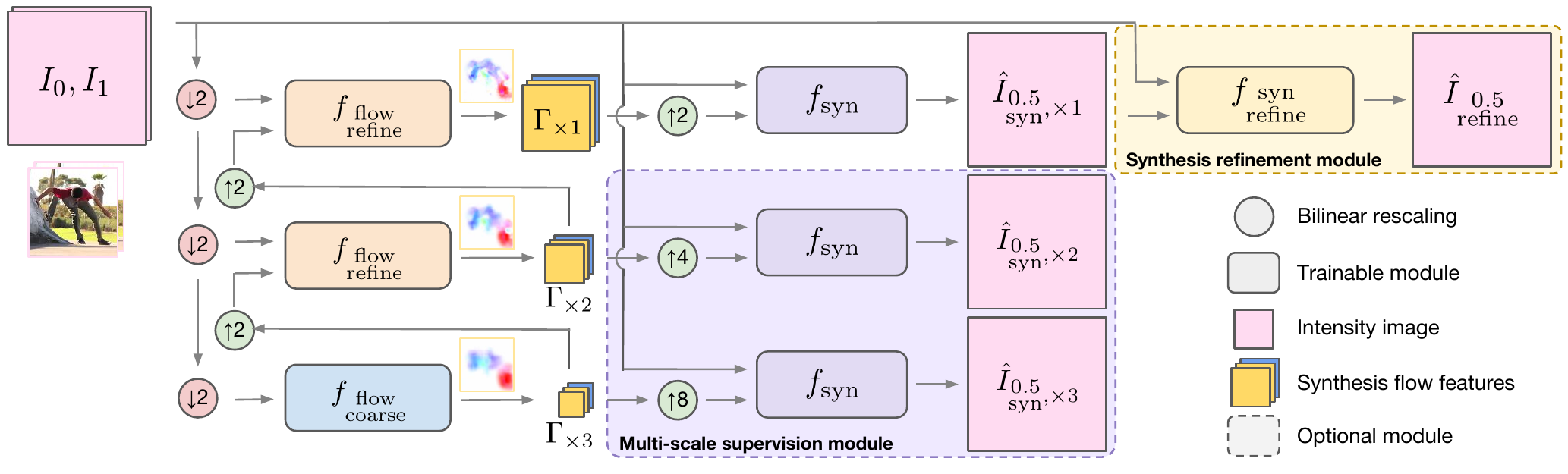}
  \caption{Overview of the frame interpolation method. Flow is estimated from 
two input frames at scales $\times 8$, $\times 4$ and $\times 2$. The finest 
flow scale is used to synthesise the final frame. Optionally, 
intermediate flow scales can be used to synthesise coarse interpolated frames in a multi-scale supervision module contributing to the training cost function, and the synthesised frame can be 
further processed through a synthesis refinement module.}
  \label{fig:highlevel-method}
\end{figure}

The method proposed is based on a trainable \gls{cnn} architecture that directly 
estimates an interpolated frame from two input frames $I_0$ and $I_1$. This 
approach is similar to the one in \cite{liu2017video}, where given many examples 
of triplets of consecutive frames, we solve an optimisation task minimising a 
loss between the estimated frame $\hat I_{0.5}$ and the ground truth 
intermediate frame $I_{0.5}$. A high-level overview of the method is illustrated 
in \cref{fig:highlevel-method}, and details about it's design and training are 
detailed in the following sections.

\subsection{Network design}
\label{sec:networkdesign}

\subsubsection{Multi-scale frame synthesis}
\label{sec:framesynthesis}

Let us assume $\Delta$ to represent the flow from time point $0.5$ to $1$, and 
for convenience, let us refer to synthesis features as $\Gamma=\{\Delta, W\}$, 
where spatial weights $W_{i,j} = [0, 1] \; \forall i, j$ can be used to handle 
occlusions and disocclusions. The synthesis interpolated frame is then given by
\begin{equation}
\hat I_{\substack{0.5 \\ \text{syn}}} = f_{\text{syn}}(I_0, I_1, \Gamma) = W 
\circ I_{0}(-\Delta) + (1-W) \circ I_{1}(\Delta),
\end{equation}
with $\circ$ denoting the Hadamard product. This is used in \cite{liu2017video} 
to synthesise the final interpolated frame and is referred to as voxel flow. 
Although a multi-scale estimation of synthesis features $\Gamma$ is presented to 
process input data at different scales, coarser flow levels are not leveraged 
for the estimation of finer flow results. In contrast, we propose to reuse a 
coarse flow estimation for further processing with residual modules, in the same spirit as in \cite{ganin2016deepwarp}.

\begin{figure}
\centering
  \includegraphics[width=0.65\columnwidth]{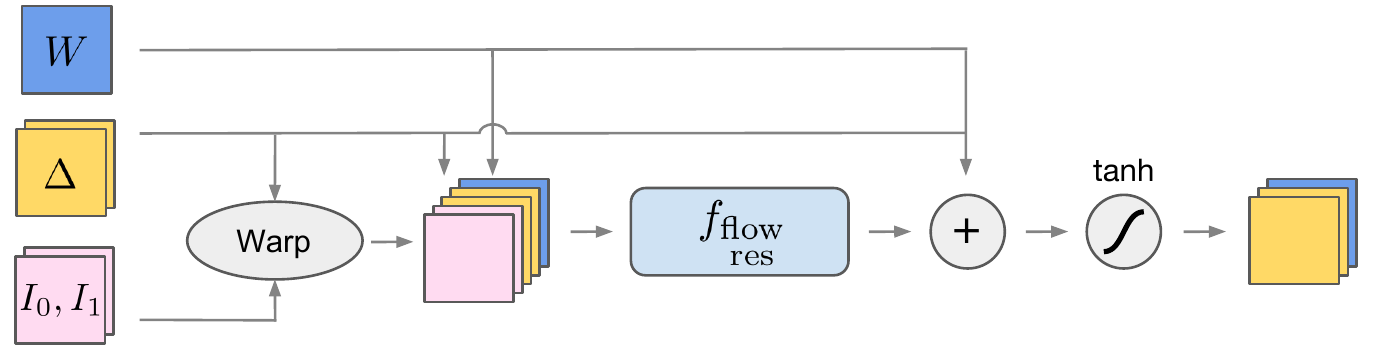}
  \caption{Flow refinement module. A coarse flow estimation wraps the 
input frames, which are then passed together with the coarse synthesis flow features to a flow residual module. The sum of the residual flow and the coarse flow features results in a fine flow estimation . A tanh non-linearity clips the result to within $[-1,1]$.}
  \label{fig:flow-refinement}
\end{figure}

To estimate synthesis features $\Gamma$ we build a pyramidal structure 
progressively applying and estimating optical flow between two frames at different 
scales $j=[1,J]$, with $J$ the coarsest level. We refer to synthesis features at different 
scales as $\Gamma_{\times j}$. If $U$ and $D$ denote $\times 2$ bilinear up- and 
down-sampling matrix operators, flow features are obtained as
\begin{equation}\label{eq:flow-highlevel}
\Gamma_{\times j} =
\begin{cases} 
      f_{\substack{\text{flow} \\ \text{coarse}}} (D^j I_0, D^j I_1) & \text{if $j=J$}, \\
      f_{\substack{\text{flow} \\ \text{refine}}} (D^j I_0, D^j I_1, U \Gamma_{\times (j+1)}) & 
\text{otherwise}.
   \end{cases}
\end{equation}
The processing for flow refinement is show in \cref{fig:flow-refinement}, and is 
formally given by
\begin{equation}
f_{\substack{\text{flow} \\ \text{refine}}}(I_0,I_1,\Gamma) = \text{tanh} (\Gamma + \Gamma_{\text{res}}),
\end{equation}
\begin{equation}
\Gamma_{\text{res}} = f_{\substack{\text{flow} \\ \text{res}}}(I_0(-\Delta), I_1(\Delta), \Gamma),
\end{equation}
with the tanh non-linearity keeping the synthesis flow features within the range 
$[-1,1]$. The coarse flow estimation and flow residual modules, $f_{\substack{\text{flow} \\ \text{coarse}}}$ and $f_{\substack{\text{flow} \\ \text{res}}}$ in \cref{fig:highlevel-method} and \cref{fig:flow-refinement} respectively, are both based on the 
\gls{cnn} architecture described in \cref{tab:conv-block}. Both modules use 
$\phi=\text{tanh}$ to produce $N_o=3$ output synthesis features within the range 
$[-1,1]$, corresponding to flow features $\Gamma$. For $3$ image color channels, coarse 
flow estimation uses $N_i=6$, and residual flow uses $N_i=9$.

Fixing $J=3$, found to be a good compromise between efficiency and performance, the final features 
are upsampled from the first scale to be $\Gamma = U\Gamma_{\times 1}$. Note 
that intermediate synthesis features can be used to obtain intermediate 
synthesis results as
\begin{equation}
\hat I_{\substack{0.5 \\ \text{syn}}, \times j} = f_{\text{syn}}(I_0, I_1, 
U^j\Gamma_{\times j}).
\end{equation}
In \cref{sec:multiscale-training} we describe how intermediate synthesis results 
can be used in a multi-scale supervision module to facilitate network training.

\subsubsection{Synthesis refinement module}
\label{sec:refinement}

Frame synthesis can be challenging in cases of complex motion or occlusions where flow estimation may be inaccurate. In these 
situations, artifacts usually produce an unnatural look for moving regions of 
the image that benefit from further correction. We therefore introduce a 
synthesis refinement module that consists of a \gls{cnn} allowing for further 
joint processing of the synthesised image with the original input frames that 
produced it. 
\begin{equation}
\hat I_{\substack{0.5 \\ \text{refine}}} = f_{\substack{\text{syn} \\ \text{refine}}} (\hat I_{\substack{0.5 \\ 
\text{syn}}}, I_0, I_1).
\end{equation}
This was shown in \cite{ganin2016deepwarp} to be beneficial in refining the 
brightness of a reconstruction result and to handle difficult occlusions. 
This module also uses the convolutional block in \cref{tab:conv-block} with 
$N_i=9$, $N_o=3$ and $\phi$ the identity function.

\begin{table}[t]
\begin{center}
\small
\begin{tabular}{c|c|c}
Layer   & Convolution kernel                & Non-linearity \\ 
\hline\hline
$1$       & $N_i \times 32 \times 3 \times 3$ & ReLU          \\
$2$, ..., $5$ & $32 \times 32 \times 3 \times 3$  & ReLU          \\
$6$       & $32 \times N_o \times 3 \times 3$ & $\phi$         
\end{tabular}
\end{center}
\caption{Convolutional network block used for coarse flow, flow residual and 
reconstruction refinement modules. Convolution kernels correspond to number of 
input and output features, followed by kernel size dimensions $3\times 3$.}
\label{tab:conv-block}
\end{table}

\subsection{Network training}
\label{sec:networktraining}

Given loss functions $l_i$ between the network 
output and the ground-truth frame, defined for an arbitrary number of components $i=\{1,I\}$, we solve
\begin{equation}\label{eq:global-problem}
\hat I_{0.5} = \argmin_{\theta} \frac{1}{N}\sum_{n=1}^N \sum_{i=1}^I \lambda_i l_i 
\left(f_{\theta}(I^n_0,I^n_1), I^n_{0.5}\right) .
\end{equation}
The output $\hat I_{0.5}$ is $\hat I_{\substack{0.5 \\ \text{syn}}}$ 
or $\hat I_{\substack{0.5 \\ \text{refine}}}$ depending on whether the refinement module is used, and $\theta$ represents all trainable parameters in 
the interpolation network $f_\theta$.

\subsubsection{Multi-scale synthesis supervision}
\label{sec:multiscale-training}

The multi-scale frame synthesis described in \cref{sec:framesynthesis} can be 
used to define a loss function at the finest synthesis scale
with a synthesis loss
\begin{equation}\label{eq:synth-scale1}
l_{\text{syn},\times 1} = \tau(\hat I_{\substack{0.5 \\ 
\text{syn}}, \times 1}, I_{0.5}),
\end{equation}
with $\tau$ a distance metric. However, an optimisation task based 
solely on this cost function suffers from the fact that it leads to an ill-posed 
solution, that is, for one particular flow map there will be multiple possible 
solutions. It is therefore likely that the solution space contains many local 
minima, making it challenging to solve via gradient descent. The network can for 
instance easily get stuck in degenerate solutions where a case with no motion, 
$\Gamma_{\times 1}=\mathbf{0}$, is represented by the network as 
$\Gamma_{\times1} = U\Gamma_{\times2} + U^2\Gamma_{\times3}$, in which case 
there are infinite solutions for the flow fields at each scale.

In order to prevent this, we supervise the solution of all scales such that flow 
estimation is required to be accurate at all scales. In practice, we define the 
following multi-scale synthesis loss function
\begin{equation}
l_{\substack{\text{multi} \\ \text{syn}}} = \sum_{j=1}^J 
\lambda_{\text{syn},j}l_{\text{syn},\times j}.
\end{equation}
We heuristically choose the weighting of this multiscale loss to be 
$\lambda_{\text{syn},j} = \{1 \text{ if } j=1; 0.5 \text{ otherwise}\}$ to 
prioritise the synthesis at the finest scale.

Additionally, the network using the synthesis refinement module 
adds a term to the cost function expressed as 
\begin{equation}\label{eq:synth-refine}
l_{\substack{\text{syn} \\ \text{refine}}} = \tau (\hat I_{\substack{0.5 \\ \text{refine}}}, 
I_{0.5}),
\end{equation}
and the total loss function for $J=3$ scales is
\begin{equation}\label{eq:total-loss}
L = l_{\substack{\text{multi} \\ \text{syn}}} + l_{\substack{\text{syn} \\ \text{refine}}} = 
l_{\text{syn},\times 1} + \frac{1}{2} [ l_{\text{syn},\times 2} + l_{\text{syn},\times 3} ] + l_{\substack{\text{syn} \\ \text{refine}}}.
\end{equation}

We propose to combine traditional pixel-wise distance metrics with higher order
metrics given by a deep network, which have been shown to correlate better with
human perception. As a pixel-wise metric we choose the $l1$-norm, which has been
shown to produce sharper interpolation results than \gls{mse}
\cite{sun2010secrets}, and we employ features 5\textunderscore 4 from the VGG network
\cite{simonyan2014very} as a perceptual loss, as proposed in
\cite{johnson2016perceptual,ledig2016photo}. Denoting with $\gamma$ the
transformation of an image to VGG space, the distance metric is therefore given by
\begin{equation}\label{eq:dist-vgg}
\tau(a,b) = \| a - b \|_1 + \lambda_{\text{VGG}} \| \gamma(a) - \gamma(b) \|^2_2.
\end{equation}
Throughout this work we fix $\lambda_{\text{VGG}}=0.001$. We include results when this term is not included in training ($\lambda_{\text{VGG}} = 0$) to analyse its impact.

\subsubsection{Generative adversarial training}
\label{sec:gan-loss}

In the loss functions described above, there is no mechanism to avoid solutions 
that may not be visually pleasing. A successful approach to force the solution 
manifold to correspond with images of a realistic appearance has been \gls{gan} 
training. We can incorporate such loss term to the objective 
function \cref{eq:total-loss} as follows

\begin{equation}\label{eq:loss-gan}
L = l_{\substack{\text{multi} \\ \text{syn}}} + l_{\substack{\text{syn} \\ \text{refine}}} + 0.0001 
l_\text{GAN}
\end{equation}

Let us call the interpolation network the \textit{generator} network 
$f_{\theta_G}$, the \gls{gan} term $l_\text{GAN}$ optimises the loss function
\begin{align}
 \min_{\theta_G}\max_{\theta_D} &\mathbb{E}_{I_{0.5} \sim 
p_{\text{train}}(I_{0.5})}[\log d_{\theta_D}(I_{0.5})] + \\
 &\mathbb{E}_{(I_0, I_1) \sim p_{f}(I_0, I_1)}[\log(1 - 
d_{\theta_D}(f_{\theta_G}(I_0, I_1))))],
\end{align}
with $d_{\theta_D}$ representing a \textit{discriminator} network that tries to 
discriminate original frames from interpolated results. The weighting parameter 
$0.0001$ was chosen heuristically in order to avoid the \gls{gan} loss 
overwhelming the total loss.

Adding this objective to the loss forces the generator $f_{\theta_G}$ to attempt 
fooling the discriminator. It has been
shown that in practice this leads to image reconstructions that incorporate 
visual properties of photo-realistic images, such as improved sharpness and 
textures \cite{ledig2016photo,radford2015unsupervised}. The discriminator 
architecture is based on the one described
in figure 4 of \cite{ledig2016photo}, with minor modifications. We start
with 32 filters and follow up with 8 blocks of convolution, batch normalization 
and leaky
ReLU with alternating strides of 2 and 1. At each block of stride 2 the number 
of features is doubled, which we found to improve the performance of the 
discriminator.

\section{Experiments}
\label{experiments}

We first compute the performance of a baseline 
version of the model that performs single-scale frame synthesis without 
synthesis refinement and is trained using a simple colour constancy $l1$-norm 
loss. We gradually incorporate to a baseline network the design and training choices proposed in \cref{sec:networkdesign} and \cref{sec:networktraining}, and evaluate their benefits visually and 
quantitatively. As a performance metric for reconstruction accuracy we use \gls{psnr}, however we note that this metric is known to not correlate well with human perception \cite{ledig2016photo}. 

\subsection{Data and implementation details}

The dataset used is a collection of 60fps videos from YouTube-8m 
\cite{abu2016youtube} resized to 
$640\times 360$. Training samples are obtained by extracting one triplet of 
consecutive frames every second, discarding samples for which two consecutive 
frames were found to be almost identical with a small squared error threshold. 
Unless otherwise stated, all models used $20$k, $1.5$k and $375$ triplets of frames 
corresponding to the training, validation and testing sets.

All network layers from convolutional building blocks based on 
\cref{tab:conv-block} are orthogonally initialised with a gain of $\sqrt{2}$, 
except for the final layer which is initialised from a normal distribution with 
standard deviation $0.01$. This forces the network to initialise flow estimation 
close to zero, which leads to more stable training. Training was performed on 
batches of size $128$ using frame crops of size $128 \times 128$ to diversify 
the 
content of batches and increase convergence speed. Furthermore, we used Adam 
optimisation \cite{kingma2014adam} with learning rate $0.0001$ and applied early 
stopping with a patience of 10 epochs. All models converge around roughly $100$ 
epochs with this setup.

\subsection{Complexity and speed analysis}

To remain framework and hardware agnostic, we report computational complexity of \glspl{cnn} in floating point operations (FLOPs) necessary for the processing of one $360 \times 640$ frame, and in the number of trainable parameters. The bottleneck of the computation is in convolutional operations to estimate a flow field and refine the interpolation, therefore we ignore operations necessary for intermediate warping stages. Using $H$ and $W$ to denote height and width, $n_l$ the number of features in layer $l$, and $k$ the kernel size, the number of FLOPs per convolution are approximated as
\begin{equation}
H W n_{l+1} \left[ 2 n_l k^2 + 2 \right].
\end{equation}

We additionally report GPU runtimes for the methods proposed as well as for SepConv $\mathcal{L}_F$ as a reference for network efficiency. Experiments were run on an NVIDIA M40 GPU.

\subsection{Network design experiments}

In the first set of experiments we evaluate the benefits of exploiting an 
implicit estimation of optical flow as well as a synthesis refinement module. 

\subsubsection{Implicit optical flow estimation}

\Glspl{cnn} can spatially 
transform information through convolutions without the need for spatial transformer pixel 
regridding. However, computing an internal 
representation of optical flow that is used by a transformer is a more efficient 
alternative to achieve this goal. In \cref{table:design} we compare results for our baseline 
architecture using multi-scale synthesis (MS $l_{\text{syn}, \times 1}$), relative to a simple \gls{cnn} that attempts to directly estimate frame $I_{0.5}$ from 
inputs $I_0$ and $I_1$. Both models are trained with $l1$-norm colour constancy loss (ie. $\lambda_{\text{VGG}}=0$ in \cref{eq:dist-vgg}). In order to replicate hyperparameters, all layers 
in the baseline \gls{cnn} model are convolutional layers of $32$ kernels of
size $3\times 3$ followed by ReLU activations, except for the last layer which 
uses a linear activation. 

\begin{figure}[t]
\centering
\includegraphics[width=0.4\linewidth]{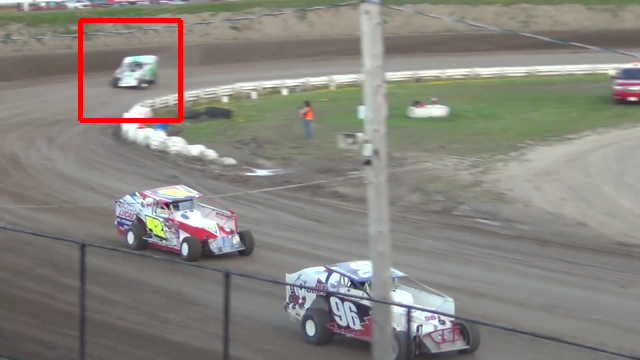}
\includegraphics[width=0.4\linewidth]{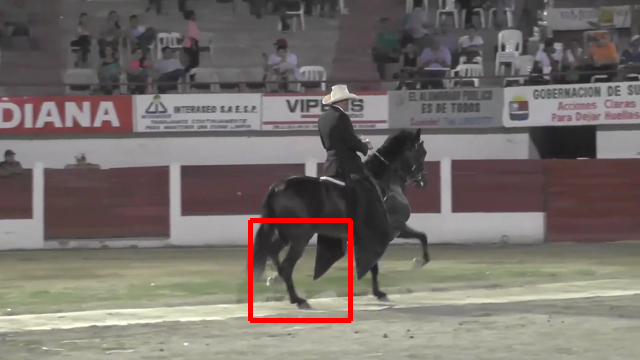}

  \begin{subfigure}{0.195\linewidth}
  \centering
  		\includegraphics[width=\linewidth]{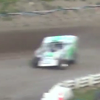}
       \caption{}
  \end{subfigure}
  \begin{subfigure}{0.195\linewidth}
  \centering
  		\includegraphics[width=\linewidth]{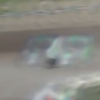}
       \caption{}
  \end{subfigure}
  \begin{subfigure}{0.195\linewidth}
  \centering
  		\includegraphics[width=\linewidth]{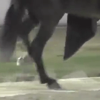}
       \caption{}
  \end{subfigure}
  \begin{subfigure}{0.195\linewidth}
  \centering
  		\includegraphics[width=\linewidth]{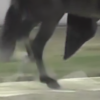}
       \caption{}
  \end{subfigure}
  
  \begin{subfigure}{0.195\linewidth}
  \centering
  		\includegraphics[width=\linewidth]{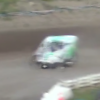}
       \caption{}
  \end{subfigure}
  \begin{subfigure}{0.195\linewidth}
  \centering
  		\includegraphics[width=\linewidth]{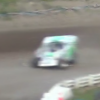}
       \caption{}
  \end{subfigure}
  \begin{subfigure}{0.195\linewidth}
  \centering
  		\includegraphics[width=\linewidth]{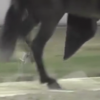}
       \caption{}
  \end{subfigure}
  \begin{subfigure}{0.195\linewidth}
  \centering
  		\includegraphics[width=\linewidth]{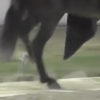}
       \caption{}
  \end{subfigure}

\caption{Impact of network design on two visual examples. Top: two full size original $360\times 640$ images with highlighted crops. Bottom: (a, c) ground-truth, (b, d) baseline CNN, (e, g) MS $l_{\text{syn}, \times 1}$, (f, h) MS $l_{\text{syn}, \times 1} + l_\text{refine}$.}
\label{fig:design}
\end{figure}

The baseline model uses $15$ layers, which results in approximately the same number of trainable parameters to the proposed spatial transformer method. Note that the multi-scale design allows to obtain an estimation with $\times 9.2$ fewer FLOPs. The baseline \gls{cnn} produces a \gls{psnr} $2.4$dB lower than multi-scale synthesis on the test set. The visualisations in \cref{fig:design} show that the baseline \gls{cnn} struggles to produce a satisfactory interpolation (b, d), and tends to produce an average of previous and past frames. The proposed multi-scale synthesis method results in more accurate approximations (e, g).

\begin{table}[b]
\begin{center}
\small
\begin{tabular}{l|lll}
Method & \gls{psnr} & Parameters & FLOPs \\
\hline\hline
Baseline CNN       & $33.93$  & $123$k & $57$G \\
MS $l_{\text{syn}, \times 1}$ & $36.31$ & $121$k & $6.1$G \\
MS $l_{\text{syn}, \times 1} + l_\text{refine}$ & $36.78$ & $161$k & $25$G \\
\end{tabular}
\end{center}
\caption{Impact of network design on performance.}
\label{table:design}
\end{table}

\subsubsection{Synthesis refinement}

Frames directly synthesised from flow 
estimation can exhibit spatial distortions leading to visually unsatisfying 
results. This limitation can be substantially alleviated with the refinement 
module described in \cref{sec:refinement}. In \cref{table:design} we also include results for a multi-scale synthesis model that additionally uses a synthesis refinement module (MS $l_{\text{syn}, \times 1} + l_\text{refine}$). This increases the number of trainable parameters by $\times 1.33$ and the number of FLOP by $\times 4.1$, but achieves adds $0.47$dB in \gls{psnr} and is able to correct inaccuracies in the estimation from the simpler MS $l_{\text{syn}, \times 1}$, as shown in \cref{fig:design} (f, h).

\subsection{Network training experiments}

In this section we analyse the impact on interpolation results brought by 
multi-scale synthesis supervision and by the use of a perceptual loss term and 
\gls{gan} training for an improved visual quality.

\subsubsection{Multi-scale synthesis supervision}

As described previously, the performance of synthesis models presented in \cref{table:design} is limited by the fact that flow estimation in 
multiple scales is ill-posed. We retrained model MS $l_{\text{syn}, \times 1} + l_\text{refine}$, showing the best performance from the design choices proposed, but changed the objective function to supervise frame interpolation at all scales as proposed in \cref{sec:multiscale-training}. This model, which we refer to as MS for brevity (short for MS $l_{\substack{\text{multi} \\ \text{syn}}} + l_\text{refine}$), increases \gls{psnr} on the test set compared to MS $l_{\text{syn}, \times 1} + l_\text{refine}$ by $0.19$dB up to $36.97$dB as shown in \cref{table:soa} when trained on the same set of $20$k training frames.

\begin{figure}[t]
\centering
\includegraphics[width=0.4\linewidth]{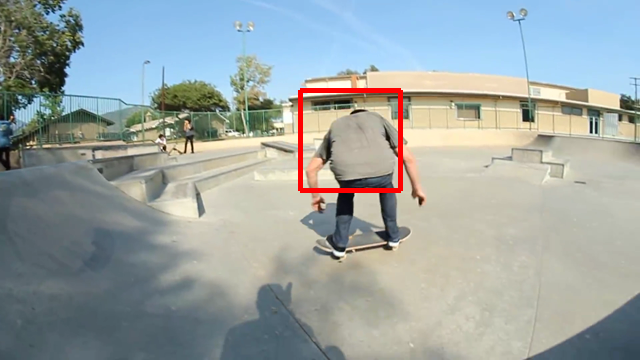}
\includegraphics[width=0.4\linewidth]{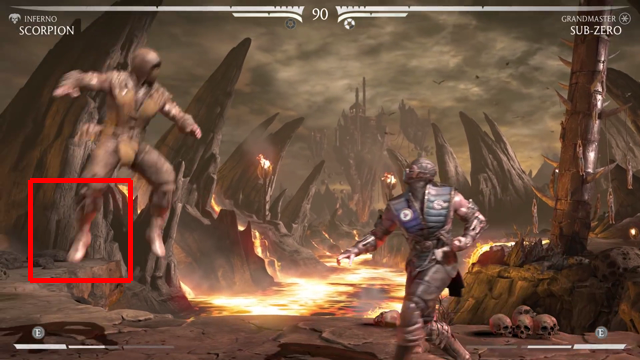}

  \begin{subfigure}{0.195\linewidth}
  \centering
  		\includegraphics[width=\linewidth]{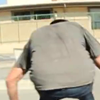}
       \caption{}
  \end{subfigure}
  \begin{subfigure}{0.195\linewidth}
  \centering
  		\includegraphics[width=\linewidth]{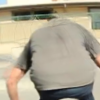}
       \caption{}
  \end{subfigure}
  \begin{subfigure}{0.195\linewidth}
  \centering
  		\includegraphics[width=\linewidth]{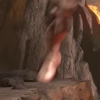}
       \caption{}
  \end{subfigure}
  \begin{subfigure}{0.195\linewidth}
  \centering
  		\includegraphics[width=\linewidth]{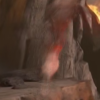}
       \caption{}
  \end{subfigure}
  
  \begin{subfigure}{0.195\linewidth}
  \centering
  		\includegraphics[width=\linewidth]{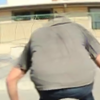}
       \caption{}
  \end{subfigure}
  \begin{subfigure}{0.195\linewidth}
  \centering
  		\includegraphics[width=\linewidth]{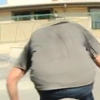}
       \caption{}
  \end{subfigure}
  \begin{subfigure}{0.195\linewidth}
  \centering
  		\includegraphics[width=\linewidth]{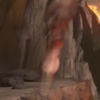}
       \caption{}
  \end{subfigure}
  \begin{subfigure}{0.195\linewidth}
  \centering
  		\includegraphics[width=\linewidth]{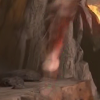}
       \caption{}
  \end{subfigure}
  
\caption{Impact of network training on two visual examples. Top: two full size original $360\times 640$ images with highlighted crops. Bottom: (a, c) ground-truth, (b, d) MS, (e, g) MS$+$VGG, (f, h) MS$+$VGG$+$GAN (FIGAN).}
\label{fig:training}
\end{figure}

\subsubsection{Impact of training data}

Unsupervised motion learning is challenging due to the large space of possible video motion. In order to learn a generalisable representation of motion, it is important to have a diverse training set with enough representative examples of expected flow directions and amplitudes. We evaluated the same mode MS when trained on different training set sizes, in particular reducing the training set to $5$k random frames and increasing it to $200$k. Although increasing the training set size inevitably increases training time, it also has a considerable impact on \gls{psnr} as shown in \cref{table:soa}. The remaining experiments use a training set size of $20$k as a compromise for performance and ease of experimentation.

\subsubsection{Perceptual loss and GAN training}

Extending the objective loss function with more abstract components such as a VGG term ($\lambda_{\text{VGG}} = 0.001$ in \cref{eq:dist-vgg}) and a GAN training strategy (\cref{eq:loss-gan}) also affects results. In \cref{table:soa} we also include results for a network MS$+$VGG, trained with the combination of $l1$-norm and VGG terms suggested in \cref{eq:dist-vgg}. We also show results for MS$+$VGG$+$GAN, which is a network additionally using adversarial training. This result of \gls{psnr} on the full test set shows that both of these modifications lower the performance relative to the simpler colour constancy training loss. However, a visual inspection of results in \cref{fig:training} demonstrate how these changes help obtaining a sharper, more pleasing interpolation. This is in line with the findings from \cite{ledig2016photo,niklaus2017sepconv}.

\begin{table}[t]
\begin{center}
\small
\begin{tabular}{l|c|c|c|c}
Method & \shortstack{Training \\ set size} & \gls{psnr} (dB) & FLOPs (G) & GPU speed (s) \\ 
\hline\hline
Farneback \cite{farneback2003two}        & - & $35.7$  & - & - \\
\hline
Deep Flow 2 \cite{weinzaepfel2013deepflow}               & - & $35.88$ & - & - \\
\hline
PCA-layers \cite{WulffCVPR2015}  & - & $36.3$ & - & - \\
\hline
Phase-based \cite{meyer2015phase}               & - & $35.17$ & - & - \\
\hline
FlowNet 2 \cite{ilg2016flownet} & - & $35.26$ & - & - \\
\hline
SepConv $\mathcal{L}_1$ \cite{niklaus2017sepconv} & \multirow{2}{*}{-} & $37.04$ & \multirow{2}{*}{$81$} & \multirow{2}{*}{$0.7$}\\
SepConv $\mathcal{L}_F$ \cite{niklaus2017sepconv} & & $36.86$ & & \\
\hline
\multirow{3}{*}{MS} & $5$k & $36.67$ & \multirow{3}{*}{$\mathbf{25}$} & \multirow{3}{*}{$\mathbf{0.015}$} \\
& $20$k & $36.97$ & & \\
& $200$k & $\mathbf{37.23}$ & & \\
\hline
MS$+$VGG & $20$k & $36.89$ & $\mathbf{25}$ & $\mathbf{0.015}$ \\
\hline
MS$+$VGG$+$GAN & \multirow{2}{*}{$20$k} & \multirow{2}{*}{$36.68$} & \multirow{2}{*}{$\mathbf{25}$} \\
(FIGAN) & & & & $\mathbf{0.015}$ \\
\end{tabular}
\end{center}
\caption{State-of the-art interpolation comparison.}
\label{table:soa}
\end{table}


\subsection{State-of-the-art comparison}

In this section, several frame interpolation methods are compared to the algorithm proposed. \Cref{table:soa} summarises \gls{psnr} results on the full tests set for all methods. The interpolation from flow-based methods \cite{farneback2003two,weinzaepfel2013deepflow,WulffCVPR2015,ilg2016flownet} was done as described in \cite{baker2011middlebury} using the optical flow features generated from implementations of the respective authors\footnote{KITTI-tuning parameters were used for PCA-Layers \cite{WulffCVPR2015}.}. The phase-based approach in \cite{meyer2015phase} and SepConv \cite{niklaus2017sepconv} are both able to directly generate an interpolated frame. We include results from SepConv using both a colour constancy loss ($\mathcal{L}_1$) and a perceptual loss ($\mathcal{L}_F$).

\begin{figure}[t]
\centering
\includegraphics[width=0.25\textwidth]{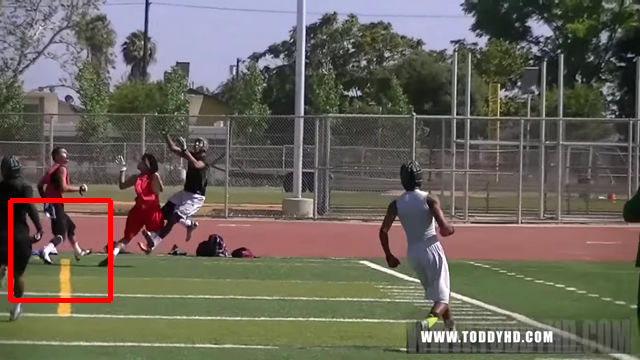}
\includegraphics[width=0.14\textwidth]{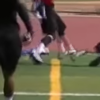}
\includegraphics[width=0.14\textwidth]{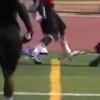}
\includegraphics[width=0.14\textwidth]{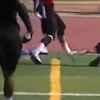}
\includegraphics[width=0.14\textwidth]{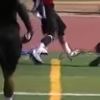}
\includegraphics[width=0.14\textwidth]{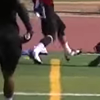}

\includegraphics[width=0.25\textwidth]{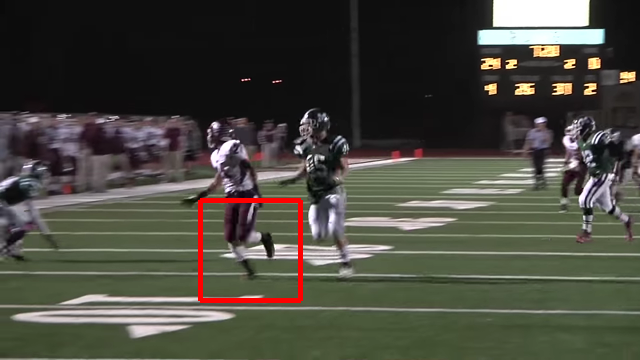}
\includegraphics[width=0.14\textwidth]{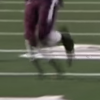}
\includegraphics[width=0.14\textwidth]{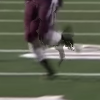}
\includegraphics[width=0.14\textwidth]{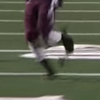}
\includegraphics[width=0.14\textwidth]{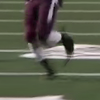}
\includegraphics[width=0.14\textwidth]{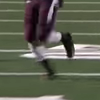}

\includegraphics[width=0.25\textwidth]{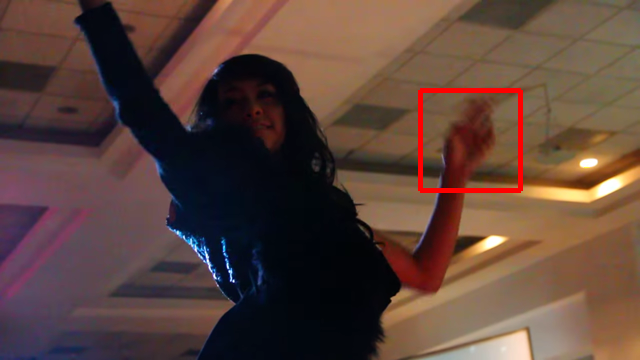}
\includegraphics[width=0.14\textwidth]{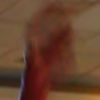}
\includegraphics[width=0.14\textwidth]{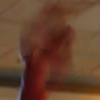}
\includegraphics[width=0.14\textwidth]{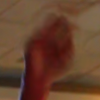} \includegraphics[width=0.14\textwidth]{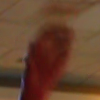}
\includegraphics[width=0.14\textwidth]{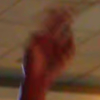}

\caption{Visualisation of sate-of-the-art comparison. From left to right: full size original $360\times 640$ image with highlighted crop, Farneback's method \cite{farneback2003two}, PCA-layers \cite{WulffCVPR2015}, SepConv $\mathcal{L}_F$ \cite{niklaus2017sepconv}, proposed FIGAN, and ground-truth.}
\label{fig:soa}
\end{figure}

As shown in \cref{table:soa}, the best performing method in terms of \gls{psnr} is MS when trained on the large training set, however we found the best visual quality to be produced by FIGAN and SepConv-$\mathcal{L}_F$, both trained using perceptual losses. Visual examples from selected methods are provided in \cref{fig:soa}. Notice that some optical flow based methods such as Farneback and PCA-layers are unable to merge information from consecutive frames correctly, which can be attributed to an inaccurate flow estimation. In contrast, FIGAN shows more precise reconstructions, and most importantly preserves sharpness and features that make interpolation results perceptually more pleasing.

\begin{figure}[t]
\centering
  \begin{subfigure}{0.22\columnwidth}
  \centering
  		\includegraphics[width=\textwidth]{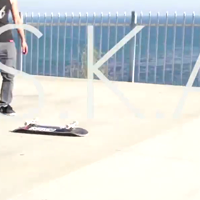}
       \caption{Original}
  \end{subfigure}
  \begin{subfigure}{0.22\columnwidth}
  \centering
  		\includegraphics[width=\textwidth]{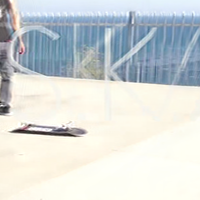}
       \caption{SepConv $\mathcal{L}_F$}
  \end{subfigure}
  \begin{subfigure}{0.22\columnwidth}
  \centering
  		\includegraphics[width=\textwidth]{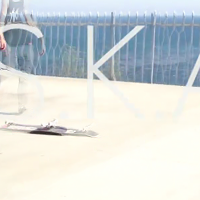}
       \caption{FIGAN}
  \end{subfigure}
\caption{SepConv $\mathcal{L}_F$ and FIGAN interpolation for conflicting overlaid motions. FIGAN favours an accurate reconstruction of the foreground while SepConv approximates the reconstruction of the background at the expense of distorting the foreground structure.}
\label{fig:fail}
\end{figure}

We found SepConv $\mathcal{L}_F$ and FIGAN to have visually comparable results for easily resolvable motion like those in \cref{fig:soa}. Their largest discrepancies in behaviour were found in challenging situations, such as static objects overlaid on top of a fast moving scene, as shown in \cref{fig:fail}. Whereas SepConv favours resolving large displacements in the background, FIGAN produces a better reconstruction of foreground objects at the expense of accuracy in the background. This could be due to the fact that SepConv estimates motion at $\times 32$ undersampling, while FIGAN only downscales by $\times 8$. Coarse-to-fine flow estimation approaches can fail when coarse scales dominate the motion of finer scales \cite{xu2012motion}, and this is likely to be more pronounced the larger the gap between the coarse and fine scales.

A relevant difference between SepConv and FIGAN is found in complexity. SepConv includes $1.81$M training parameters, which is $\times 11.2$ more than FIGAN, but also each $360 \times 640$ frame interpolation requires $81$G FLOPs, or $\times 3.24$ more compared to FIGAN. Noting a comparable visual quality and \gls{psnr} figures for a small fraction of training parameters, this highlights the efficiency advantages of FIGAN, which was designed under real-time constraints. The proposed networks run in real-time, attain state-of-the-art performance, and are $\times 47$ faster than the closest competing method.

\section{Conclusion}
\label{conclusion}
In this paper, we have described a multi-scale network based on recent advances in spatial transformers and composite perceptual losses. Our proposed architecture sets a new state of the art in terms of \gls{psnr}, and produces visual quality results comparable to the best performing neural network solution with $\times 3.24$ fewer computations. Our experiments confirm that a network design drawing from traditional pyramidal flow refinement allows to reduce its complexity while maintaining a competitive performance. Furthermore, training losses beyond classical pixel-wise metrics and adversarial training provide an abstract representation that translate into sharper, and visually more pleasing interpolation results.

\bibliographystyle{splncs}
\bibliography{egbib}
\end{document}